\documentclass[10pt,twocolumn,letterpaper]{article}

 \usepackage{cvpr}              % To produce the CAMERA-READY version
\definecolor{cvprblue}{rgb}{0.21,0.49,0.74}
\usepackage[pagebackref,breaklinks,colorlinks,allcolors=cvprblue]{hyperref}
\usepackage{amsmath}
\usepackage{bm}
\usepackage[accsupp]{axessibility}

\usepackage{amsmath,amsfonts,bm}

\def\eqref#1{equation~\ref{#1}}
\def\1{\bm{1}}

\def\rvc{{\mathbf{c}}}

\def\rvz{{\mathbf{z}}}

\def\vtheta{{\bm{\theta}}}

\def\vf{{\bm{f}}}

\DeclareMathAlphabet{\mathsfit}{\encodingdefault}{\sfdefault}{m}{sl}
\SetMathAlphabet{\mathsfit}{bold}{\encodingdefault}{\sfdefault}{bx}{n}

\newcommand{\E}{\mathbb{E}}

\usepackage{hyperref}
\usepackage{url}
\usepackage{graphicx}
\usepackage{enumitem}
\usepackage{wrapfig}
\usepackage[capitalize]{cleveref}
\usepackage{tabularx}
\usepackage{animate}
\usepackage{booktabs}
\usepackage{caption}
\usepackage{subcaption}
\usepackage{multirow}

\definecolor{deepyellow}{RGB}{255, 204, 0} 
\def\paperID{16022} % *** Enter the Paper ID here
\def\confName{CVPR}
\def\confYear{2025}

\title{MotionPro: A Precise Motion Controller for Image-to-Video Generation\thanks{{\small This work was performed at HiDream.ai.}}}

\author{\normalsize Zhongwei Zhang$^{1}$, Fuchen Long$^{2}$, Zhaofan Qiu$^{2}$, Yingwei Pan$^{2}$\thanks{Co-corresponding author.}, Wu Liu$^{1}$\footnotemark[2], Ting Yao$^{2}$,  and Tao Mei$^{2}$\\
	$^{1}$\normalsize University of Science and Technology of China \quad $^{2}$\normalsize HiDream.ai Inc. \\
	{\tt\small\ zhwzhang@mail.ustc.edu.cn}, {\tt\small\{longfuchen, qiuzhaofan, pandy\}@hidream.ai} \\
	{\tt\small\ liuwu@live.cn}, {\tt\small\{tiyao, tmei\}@hidream.ai} \\
}

\begin{document}
\maketitle

\begin{abstract}
Animating images with interactive motion control has garnered popularity for image-to-video (I2V) generation. 
Modern approaches typically rely on large Gaussian kernels to extend motion trajectories as condition without explicitly defining movement region, leading to coarse motion control and failing to disentangle object and camera moving. 
To alleviate these, we present MotionPro, a precise motion controller that novelly leverages region-wise trajectory and motion mask to regulate fine-grained motion synthesis and identify target motion category (i.e., object or camera moving), respectively.
Technically, MotionPro first estimates the flow maps on each training video via a tracking model, and then samples the region-wise trajectories to simulate inference scenario.
Instead of extending flow through large Gaussian kernels, our region-wise trajectory approach enables more precise control by directly utilizing trajectories within local regions, thereby effectively characterizing fine-grained movements.
A motion mask is simultaneously derived from the predicted flow maps to capture the holistic motion dynamics of the movement regions.
To pursue natural motion control, MotionPro further strengthens video denoising by incorporating both region-wise trajectories and motion mask through feature modulation.
More remarkably, we meticulously construct a benchmark, i.e., MC-Bench, with 1.1K user-annotated image-trajectory pairs, for the evaluation of both fine-grained and object-level I2V motion control.
Extensive experiments conducted on WebVid-10M and MC-Bench demonstrate the effectiveness of MotionPro.
Please refer to our project page for more results: \url{https://zhw-zhang.github.io/MotionPro-page/}.
\end{abstract}

\section{Introduction}
In recent years, diffusion models \cite{2022vdm, 2022imagenvideo,2023make-a-video,2023cogvideo,2023modelscope,2024animatediff,yang2024hi3d,2023photorealistic,guo2023i2v,qu2024chatvtg,liu2024hcma,long2024eccv,zhang2024trip,chen2024sateco,qian2024masf,chen2025ouroboros,wan2025tryon} have shown significant progress in revolutionizing text-to-video (T2V) generation.
Although promising visual appearance can be attained by these advances, the controllable motion generation is still a grand challenge in video diffusion paradigm.
There are several attempts \cite{2023videocomposer, 2023text2video-zero, 2023structure, 2024controlvideo} to enhance controllable capacity of video synthesis with additional guidance (e.g., depth, edge or optical flow).
Nevertheless, it might be impractical for users to conveniently provide such signals as input conditions.
Hence, the focus of this paper is to capitalize on the user-friendly conditions (i.e., sparse trajectory and region mask) for enabling interactively controllable image-to-video (I2V) generation: given the reference image as the first frame, the motion in the synthesized video should be natural and well-aligned with the provided trajectory.

\begin{figure}
	\centering
	\includegraphics[width=1.0\linewidth]{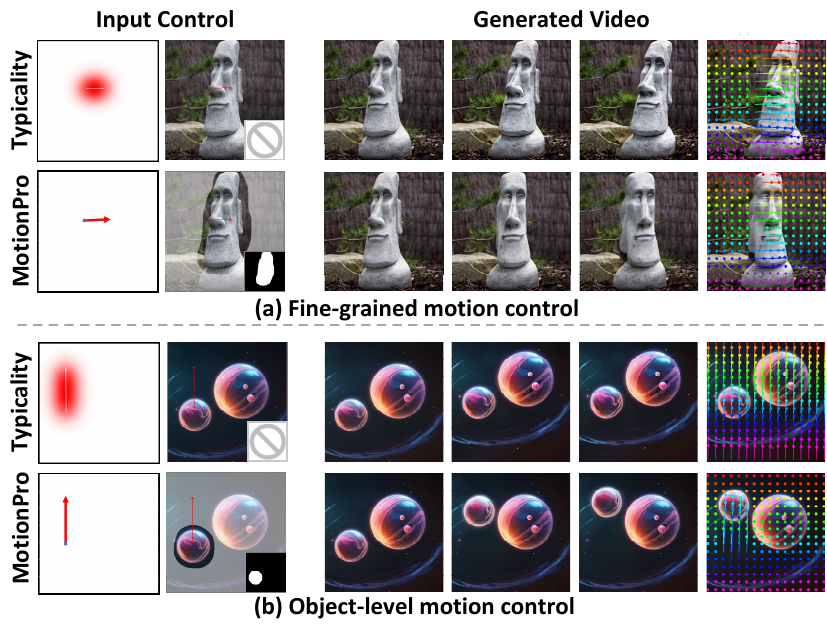}
	\vspace{-0.23in}
	\caption{An illustration of (a) fine-grained and (b) object-level motion control by using typical Gaussian filtered trajectory and our MotionPro. The flow of generated videos are also visualized.}
	\label{fig:intro}
	\vspace{-0.22in}
\end{figure}

Pioneering practices \cite{2023dragnvwa, 2024draganything} of controllable I2V generation usually guide video denoising with the single condition of Gaussian filtered trajectory.
In the training stage, the input trajectories are first sparsely sampled from the optical flow maps and then processed by a Gaussian filter with large kernel to mitigate pixel-level trajectory instability (e.g., $99$ in DragNUWA~\cite{2023dragnvwa}).
%The flow approximation brought by Gaussian filtering inevitably results in the inaccuracy of fine-grained motion details and limits the model capability for precise motion control.
The flow extension brought by Gaussian filtering inevitably results in the inaccuracy of fine-grained motion details and limits the model capability for precise motion control.
Therefore, the generated fine-grained movement (e.g., the turning-head of first case in Figure~\ref{fig:intro}) is unnatural.
Another issue is that the single condition of trajectory commonly fails to precisely identify the target motion category (i.e., object or camera moving).
For instance, as depicted in Figure~\ref{fig:intro}, the trajectory on the planet could be explained as two moving situations, i.e., the camera being pulled downwards with relative to static two planets (camera movement) or planet rising corresponding to static background (object movement).
Solely relying on the trajectory might lead to the motion misinterpretation and thus hinder exactly controllable I2V generation.
To address the above two issues, we shape a new paradigm of motion controller that capitalizes on region-level trajectory and motion mask to enhance video denoising for controllable I2V synthesis.
Specifically, we spatially sample multiple local regions in the video optical flow maps and directly employ the trajectories in the sparse regions as input condition.
In this way, the attained trajectories could maintain accurate motion details and enables adequate capture of fine-grained motion.
Meanwhile, a motion mask is estimated on the optical flow maps which aims to globally emphasize the motion area, thereby specifying the target motion category and alleviating misinterpretation.
To further regulate motion synthesis, we predict the affine parameters on the collaboration of trajectory and motion mask to modulate the video latent codes in denoising.
As shown in Figure~\ref{fig:intro}, our unique region-wise trajectory design and the employment of motion mask achieves the better fine-grained (e.g., turning-head) and object-level (e.g., planet-moving) motion.

By materializing the idea of facilitating controllable I2V generation with the proposed conditions, we present a novel framework, namely MotionPro, a precise region-wise motion control.
Specifically, given the input video, MotionPro first estimates the sequence of visibility masks and optical flow maps by using an off-the-shelf optical tracking model.
Next, the global visibility mask is obtained through computing the intersection of all visibility masks, and further multiplied with the flow map of each frame.
Then, MotionPro splits the masked flow maps into multiple local regions (e.g., the region with the size of $8\times8$) and employs the trajectories on such sparsely-sampled regions as region-wise trajectory.
Meanwhile, MotionPro attains the motion mask on the flow maps via thresholding mechanism for representing holistic motion.
Given the region-wise trajectory and corresponding motion mask, the multi-scale features are learnt by a motion encoder, and further employed to predict scale and bias for video latent feature modulation. 
Moreover, MotionPro fine-tunes all attention modules in 3D-UNet via utilizing the Low-Rank Adaptation (LoRA) technique to pursue better motion-trajectory alignment. 

The main contribution of this work is design of MotionPro by leveraging region-wise trajectory and motion mask as the complementary control signals for precise controllable I2V diffusion.
Beyond this, one benchmark, i.e., MC-Bench, with 1.1K user-annotated image-trajectory pairs, is carefully collected for evaluation.
Extensive experiments further verify the superiority of MotionPro in terms of both video quality and motion-trajectory alignment.

\section{Related Work}
\textbf{Image-to-Video Diffusion Models.}
The remarkable progress achieved by text-to-video generation~\cite{2022vdm, 2022imagenvideo,2023align,2023make-a-video,2023cogvideo,2023pyoco,2023modelscope,2024animatediff,2023photorealistic,2024lumiere,sora2024,2024cogvideox, 2023text2video-zero, 2023videofusion,2024vidu,2024latte} encourages the development of image-to-video (I2V) diffusion models.
These advances~\cite{2023emu,svd2023,2023videocomposer,2024dynamicrafter,2023seine,2023i2vgen,2024motion,2024make} treat static image as the input condition for temporal coherent video synthesis.
VideoComposer~\cite{2023videocomposer} is one of the earlier works that integrates image condition into 3D-UNet through concatenating the clean image latent with the noisy video latents.
Based on this recipe, DynamiCrafter~\cite{2024dynamicrafter} and SVD~\cite{svd2023} additionally inject the CLIP~\cite{2021clip} feature of reference image into video denoising to enhance the information guidance.
To achieve high-resolution I2V generation, I2VGen-XL~\cite{2023i2vgen} introduces a cascading diffusion model to first animate image in the low resolution and further magnifies it via video refinement. 
Besides, there are several explorations~\cite{2023seine,2024make} that simultaneously utilize two images (i.e., the first and last frames) as more powerful references to elevate I2V generation.
In this work, we choose the pre-trained I2V diffusion model SVD~\cite{svd2023} as our base architecture for motion control.

\begin{figure*}
	\centering
	\includegraphics[width=0.95\linewidth]{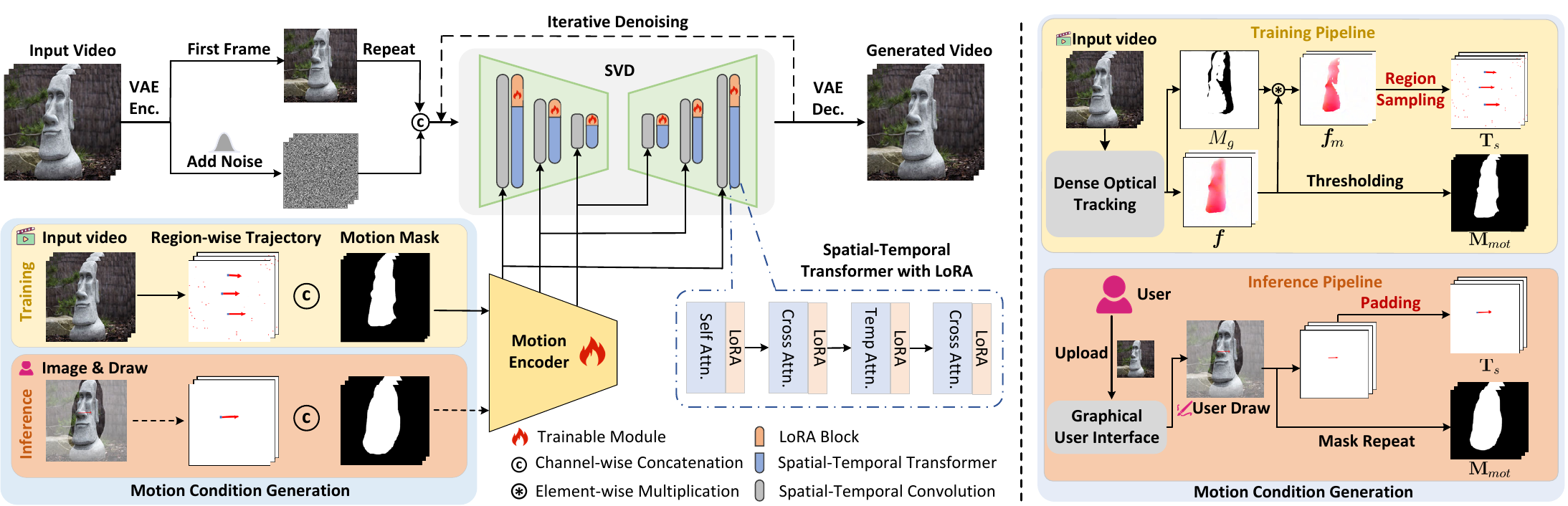}
	\vspace{-0.11in}
	\caption{An overview of (a) our MotionPro for controllable I2V generation and (b) pipeline of motion condition generation. During training, MotionPro first extracts the proposed region-wise trajectory and motion mask on the input video as the control signals. The multi-scale features are then learnt on these signals by a motion encoder, and further injected into the 3D-UNet of SVD in a feature modulation manner. Meanwhile, LoRA layers are integrated into all attention modules in the transformer blocks to improve the optimization of motion-trajectory alignment. In the inference stage, the region-wise trajectory and motion mask are first derived from the user provided trajectory and brushed region, and then exploited as the guidance to calibrate I2V video generation.}
	\label{fig:framework}
	\vspace{-0.2in}
\end{figure*}

\textbf{Controllable Video Diffusion Models.}
Despite high-quality video synthesis via I2V diffusion models, the controllable motion generation still remains an under-explored problem.
The early controllable video diffusion techniques~\cite{2023videocomposer, 2023text2video-zero, 2023structure, 2023control, 2024controlvideo,2023stablevideo} typically leverage the condition of depth, edge or optical flow, for particular motion generation.
Nevertheless, it is usually impractical for users to conveniently obtain such kinds of signals.
To address this issue, the studies exploring bounding box~\cite{2024peekaboo,2024boximator,2023trailblazer,2024motionbooth, 2024direct} or trajectory~\cite{2023dragnvwa,2024mofa,2024revideo,2024motionctrl} as additional condition for motion control start to emerge. 
%One representative of using bounding box as control is PEEKABOO~\cite{2024peekaboo} which designs the training-free spatial-temporal masked attention for visual-textual alignment in the box.
In the direction of utilizing trajectory condition, DragNUWA~\cite{2023dragnvwa} and DragAnything~\cite{2024draganything} exploits Gaussian filtered trajectory to regulate motion synthesis to mitigate pixel-level trajectory instability. However, this approach limits fine-grained motion controllability and often fails to disentangle object and camera motions when relying solely on trajectories as the conditional input.
Recently, Motion-I2V~\cite{2024motion} and MOFA-Video~\cite{2024mofa} proposed a two-stage motion control framework, which first densifies input trajectories through a sparse-to-dense network and subsequently regulates video denoising using the estimated dense trajectories. 
Similarly, MOFA-Video also introduces the concept of a movement region mask. However, the mask is solely employed for flow masking as a post-processing step, rather than being integrated as a conditional input for motion-region-aware video generation. This limitation can occasionally lead to unnatural motions and localized distortions in the synthesized videos.
%Overall, previous methods that rely on large Gaussian kernels for flow extension limit the capacity for fine-grained motion control. Additionally, using only trajectories as the network input may cause the model to fail in disentangling object and camera motions or generate locally distorted movements without motion-region awareness.
Instead, our work focuses on a new recipe of leverages region-wise trajectories and motion masks for precisely controllable I2V generation. The proposal of MotionPro contributes by studying not only how to express the motion trajectory accurately, but also how to benefit natural and precise motion generation with the synergy of the region-wise trajectory and motion mask.

\section{Our Approach}
Here, we introduce our MotionPro framework for controllable I2V generation.
Figure \ref{fig:framework} illustrates an overview of our architecture.
Given a video clip at training, the newly-minted region-wise trajectory and motion mask are first extracted as the control signals.
Next, multi-scale features are learnt on the concatenation of the trajectory and mask via a motion encoder.
These features are further injected into the 3D-UNet of SVD~\cite{svd2023} to regulate video denoising.
In each feature scale of the 3D-UNet, a scale and bias are predicted through convolutional layers to modulate the feature of video latent codes.
Besides, all attention modules are fine-tuned by LoRA~\cite{2021lora} to attain better alignment between the synthesized motion and input trajectory.

\subsection{Preliminaries: Stable Video Diffusion}
To leverage comprehensive motion prior embedded in the pre-trained diffusion models for video generation, we exploit the advanced I2V generation model, i.e., Stable Video Diffusion (SVD)~\cite{svd2023} as the base architecture of our MotionPro.
To better understand our proposal, we first revisit the training procedure of SVD.
Formally, given an input video clip $\mathbf{x}_0 = \{x_0^i\}_{i=1}^{L}$ with $L$ frames, the clean video latent codes $\mathbf{z}_0 = \{z_0^i\}_{i=1}^{L}$ are first extracted via a variational auto-encoder (VAE).
Then, the Gaussian noise $\mathbf{n}$ is added to $\mathbf{z}_0$ through forward diffusion procedure as: 
\begin{equation}\label{eq:add_noise}
	\mathbf{z} = \mathbf{z}_0 + \mathbf{n},~~~(\sigma, \mathbf{n}) \sim p(\sigma, \mathbf{n}),
\end{equation}
where $\mathbf{z}$ is the noised video latent codes and $p(\sigma, \mathbf{n})=p(\sigma)\mathcal{N}(\mathbf{0}, \sigma^2\mathbf{I})$. $\sigma$ represents the noise level and $p(\sigma)$ is the pre-determined distribution over $\sigma$. 
Following the training protocol of EDM~\cite{2022EDM}, SVD leverages the 3D-UNet $F_\vtheta$ (with parameters $\vtheta$) to predict the clean video latent codes $\hat{\mathbf{z}}_0 $ with the condition of input noised latents $\mathbf{z}$, noise level $\sigma$ and the reference image $\mathbf{c}_I$:
\begin{equation}\label{eq:param_equal}
	\hat{\mathbf{z}}_0 =c_\text{skip}(\sigma)\mathbf{z}+c_\text{out}(\sigma)F_{\vtheta}(c_\text{in}(\sigma)\mathbf{z},~ \mathbf{c}_I;~c_\text{noise}(\sigma)),
\end{equation}
where $c_\text{skip}(\sigma)$, $c_\text{out}(\sigma)$, $c_\text{in}(\sigma)$ and $c_\text{noise}(\sigma)$ are pre-defined hyper-parameters determined by noise level $\sigma$.
In SVD, the information of reference frame is injected into 3D-UNet along two pathways: a) the channel-wise concatenation of noised video latent codes and first frame latent code; b) the cross-attention between video latent feature and image CLIP~\cite{2021clip} embedding of first frame.
The loss function is formulated via denoising score matching (DSM) as:
\begin{align} \label{eq:diffusion_objective}
	\mathcal{L} = \E_{\substack{\rvz_0, \rvc_I, (\sigma, \mathbf{n}) \sim p(\sigma, \mathbf{n})}} \left[\lambda_\sigma \|\hat{\mathbf{z}}_0 - \rvz_0 \|_2^2 \right],
\end{align}
where $\lambda_\sigma$ is a weighting function.
In the scenario of our work, besides the condition of reference first frame, we additionally exploit a new kind of region-wise trajectory and motion mask as the control signals to refine video denoising for motion control.

\subsection{Motion Condition Generation}
Most existing controllable I2V approaches calibrate the video denoising with the sole guidance of Gaussian filtered point-wise trajectory.
Nevertheless, the flow extending brought by Gaussian filtering may result in inaccuracy of fine-grained motion details.
Therefore, the ability of precise motion control could be limited.
Besides, solely relying on the trajectory for motion control might not exactly express target motion category (i.e., camera or object moving), leading to motion misinterpretation in video generation.
To alleviate these issues, we propose to directly sample trajectories from optical flow maps in multiple local regions as the region-wise trajectory.
Such trajectory could preserve more precise motion details and thus manage to characterize fine-grained movement.
Meanwhile, a motion mask is further derived from the flow maps to explicitly identify target motion category of the generated videos.

\textbf{Region-wise Trajectory.}
As depicted in Figure~\ref{fig:framework}, given the input video clip $\mathbf{x}_0=\{x_0^i\}_{i=1}^{L}$ with the size of $L\times H\times W\times 3$, we first employ a dense optical tracking model, i.e., DOT~\cite{2024dense} to estimate optical flow maps $\vf=\{f^i\}_{i=1}^{L}$ and the sequence of visibility masks $\mathbf{M}=\{M^i\}_{i=1}^{L}$:
\begin{equation}\label{eq: dot}
	f^i, M^i = \text{DOT}(\mathbf{x}_0^1,\mathbf{x}_0^i),~~~i=1,2,...,L,
\end{equation}
where $f^i \in \mathbb{R}^{H\times W\times 2}$ and $M^i \in \{0,1\}^{H\times W}$ is the optical flow map and the visibility mask between the first and the $i$-th frame, respectively.
Then, we calculate the intersection on $\mathbf{M}$ to attain a global visibility mask $M_g \in \{0, 1\}^{H\times W}$ that indicates the locations having visible optical flow along temporal dimension as:
\begin{equation}\label{eq: single_visible_mask}
	M_g = \prod_{i=1}^{L}{M^i}.
\end{equation}
Next, the masked flow maps $\vf_m=\{f_m^i\}_{i=1}^L$ are computed by frame-wisely multiplying the flow maps $\vf$ with the global visibility mask $M_g$ as follows:
\begin{equation}\label{eq: valid_traject}
	\vf_m = \{f^i \cdot M_g\}_{i=1}^L.
\end{equation}
We split the masked flow maps $\vf_m$ into multiple local regions and the spatial size of each region is $k\times k$. The region-wise trajectories $\mathbf{T}_s\in \mathbb{R}^{L\times H\times W\times 2}$ are finally sampled from the region-split $\vf_m$ with a region selection mask $M_{sel} \in \{0, 1\}^{\frac{H}{k}\times \frac{W}{k}}$: 
\begin{equation}\label{eq: traj}
	\mathbf{T}_s = \{f_m^i \cdot Pad(M_{sel})\}_{i=1}^L,
\end{equation}
where $M_{sel}$ is uniformly sampled from $\{0, 1\}$ with the mask ratio $r_m$, and $Pad(\cdot)$ denotes the padding function which fills the mask value into the $k\times k$ region around each position.
Instead of exploiting a constant mask ratio for trajectory selection, we randomly choose $r_m$ in a range of $[r_{min}, 1.0]$ to simulate different real-world motion masking scenarios, which benefits the robust network optimization.  
%In this way, there is no flow approximation of the trajectories in each local region, enhancing the control ability of fine-grained motion in I2V models. 
In this way, we formulate a more precise signal by exploiting the trajectories in local region, enhancing the control ability of fine-grained motion in I2V models.

\textbf{Motion Mask.}
In addition to the region-wise trajectory for video denoising regulation, the motion mask aims to specify the motion category and benefit the global motion correlation.
Given the flow maps $\vf=\{f^i\}_{i=1}^{L}$ estimated by DOT, we first calculate the average flow magnitude $f_{avg} \in \mathbb{R}^{ H\times W}$ along temporal dimension as: $f_{avg} = \frac{1}{L} \cdot \sum_{i=1}^{L}\parallel f^i\parallel_2$.
Then, we construct the motion mask $M_{mot} \in \{0,1\}^{H\times W}$ from zero matrix, and set the value of the position where $f_{avg}$ is greater than $1$ as True. 
$M_{mot}$ is finally repeated $L$ times as the motion mask sequence $\mathbf{M}_{mot} \in \{0,1\}^{L\times H\times W\times 1}$ to align the temporal length of input video for subsequent motion control learning.

\begin{figure}
	\centering
	\vspace{-0.2in}
	\includegraphics[width=0.80\linewidth]{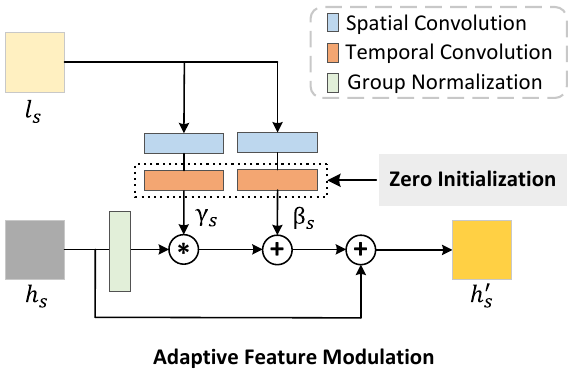}
	\vspace{-0.15in}
	\caption{An illustration of adaptive feature modulation.}
	\vspace{-0.2in}
	\label{fig:norm}
\end{figure}

\subsection{Motion Control Learning} 
With the region-wise trajectory and motion mask, we aim to control motion generation with the input signals.
Inspired by the recipe of feature adaptation in controllable image generation~\cite{2023adding}, we propose to exploit a lightweight motion encoder to estimate multi-scale features on the conditions, and utilize these features to adaptively modulate video latent feature in each corresponding scale.
To further improve the alignment between input trajectory and generated video, we fine-tune all attention modules in the spatial-temporal transformer blocks of 3D-UNet via using LoRA~\cite{2021lora}.

\textbf{Adaptive Feature Modulation.}
Given the attained region-wise trajectory $\mathbf{T}_s$ and motion mask $\mathbf{M}_{mot}$, we first concatenate them along channel dimension to form the input condition.
As shown in Figure~\ref{fig:framework}, a lightweight motion encoder with a series of convolutional layers first encodes the input condition into multi-scale feature maps.
In each scale, the learnt feature map is employed to modulate the video latent feature at the same scale in 3D-UNet.
Figure~\ref{fig:norm} depicts an illustration of the adaptive feature modulation by using the feature map $l_s$ in $s$-th scale.
Particularly, we estimate the scale $\gamma_s$ and bias $\beta_s$ on $l_s$ via a spatial-temporal convolutional layer.
Then, the normalized feature map of the input video latent feature $h_s$ is modulated via $\gamma_s$ and $\beta_s$, and further added back to itself in a skip-connection manner to form the output feature map $h_s^{\prime}$ as:
\begin{equation}\label{eq: hidden_state}
	h_s^{\prime} = GN(h_s) \cdot \gamma_s + \beta_s + h_s,
\end{equation}
where $GN(\cdot)$ denotes the group normalization.
Note that we implement zero initialization on temporal convolutional layers to initialize $\gamma_s$ and $\beta_s$ as zero at the beginning of training, guaranteeing the stability of model optimization.

\textbf{LoRA Integration.}
To preserve rich motion prior learnt by the pre-trained video diffusion model and elevate the effectiveness of motion control, we employ LoRA layers in all attention modules of spatial-temporal transformer blocks as demonstrated in Figure~\ref{fig:framework}.
Specifically, the LoRA parameters $\Delta \mathcal{W}$ act as a residue part of the original weights $\mathcal{W}$: 
\begin{equation}
	\mathcal{W}' = \mathcal{W} + \Delta \mathcal{W} = \mathcal{W} + AB^T,
\end{equation}
where $\mathcal{W}'$ is the fused weights of attention module. $A$ and $B$ are trainable matrices in LoRA layers.

\subsection{Inference Pipeline of MotionPro} 
Our MotionPro is a user-friendly I2V generation framework for interactive motion control. 
In the inference stage, as shown in Figure~\ref{fig:framework}, users can readily brush the motion region on the uploaded reference image and draw the trajectory of moving direction as input control signals. 
In detail, the motion mask can be directly obtained from the user provided brush mask.
Given the user trajectory which generally describes the movement of a single pixel, we pad the trajectory value in the $k\times k$ region around the pixel position to match the training paradigm.
The padded trajectory in local region is exploited as the input region-wise trajectory.
Finally, MotionPro regulates video denoising with the guidance of the two collaborative control signals through adaptive feature modulation. 
Both fine-grained and object-level motion control are facilitated by the synergy of the proposed region-wise trajectory and motion mask. 

\begin{figure*}[t]
	\centering
	\begin{tabularx}{0.88\textwidth}{XXXXX}
		\centering \textbf{Input Control} & \centering \textbf{DragNUWA} & \centering \textbf{DragDiffusion} & \centering \textbf{MOFA-Video} & \centering \textbf{MotionPro}
	\end{tabularx}
	\resizebox{0.88\textwidth}{!}{
		\begin{tabular}{
				@{}
				p{\dimexpr(\textwidth-0.3em*4)/5} @{\hspace{0.3em}}
				p{\dimexpr(\textwidth-0.3em*4)/5} @{\hspace{0.3em}}
				p{\dimexpr(\textwidth-0.3em*4)/5} @{\hspace{0.3em}}
				p{\dimexpr(\textwidth-0.3em*4)/5} @{\hspace{0.3em}}
				p{\dimexpr(\textwidth-0.3em*4)/5} @{}}
			\includegraphics[width=\linewidth]{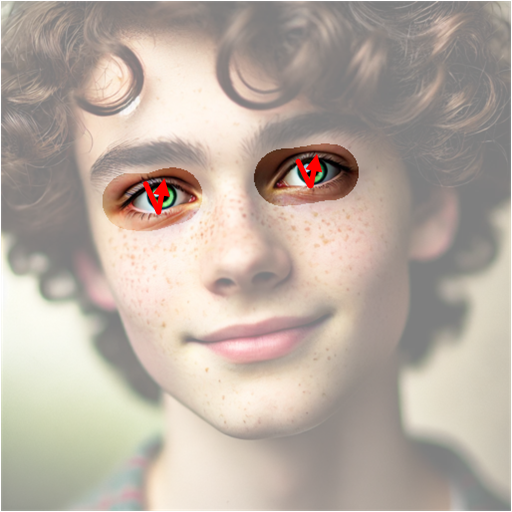} &  
			\animategraphics[autoplay,loop,width=\linewidth]{8}{videos/fined_case_final/dragnvwa/case1/pngs/}{01}{16}     & 
			\animategraphics[autoplay,loop,width=\linewidth]{8}{videos/fined_case_final/dragdiffusion/case1/pngs/}{01}{16} & 
			\animategraphics[autoplay,loop,width=\linewidth]{8}{videos/fined_case_final/mofa/case1/pngs/}{01}{16} &
			\animategraphics[autoplay,loop,width=\linewidth]{8}{videos/fined_case_final/ours/case1/pngs/}{01}{16} \\
			
			\includegraphics[width=\linewidth]{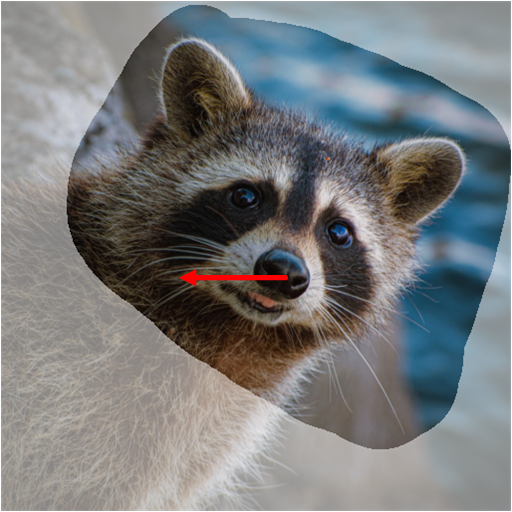} & 
			\animategraphics[autoplay,loop,width=\linewidth]{8}{videos/fined_case_final/dragnvwa/case2/pngs/}{01}{16}     & 
			\animategraphics[autoplay,loop,width=\linewidth]{8}{videos/fined_case_final/dragdiffusion/case2/pngs/}{01}{16} & 
			\animategraphics[autoplay,loop,width=\linewidth]{8}{videos/fined_case_final/mofa/case2/pngs/}{01}{16} &
			\animategraphics[autoplay,loop,width=\linewidth]{8}{videos/fined_case_final/ours/case2/pngs/}{01}{16}  
			\\
			
			\includegraphics[width=\linewidth]{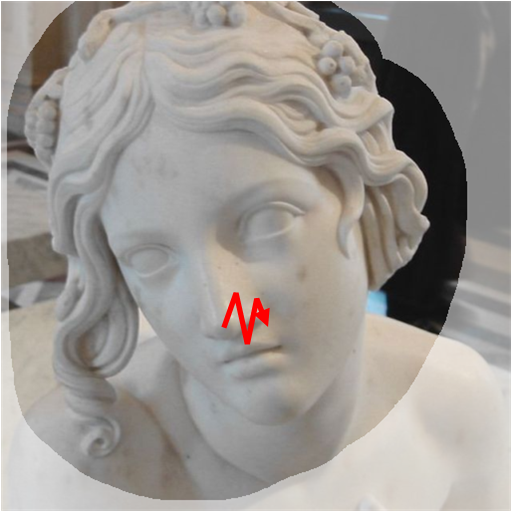} & 
			\animategraphics[autoplay,loop,width=\linewidth]{8}{videos/fined_case_final/dragnvwa/case3/pngs/}{01}{16}     & 
			\animategraphics[autoplay,loop,width=\linewidth]{8}{videos/fined_case_final/dragdiffusion/case3/pngs/}{01}{16} & 
			\animategraphics[autoplay,loop,width=\linewidth]{8}{videos/fined_case_final/mofa/case3/pngs/}{01}{16} &
			\animategraphics[autoplay,loop,width=\linewidth]{8}{videos/fined_case_final/ours/case3/pngs/}{01}{16} 
	\end{tabular}}	
	\vspace{-0.1in}
	\caption{Examples of fine-grained motion control results on MC-Bench. The input control signals include the reference image, trajectory and motion mask. \textit{\textbf{Best viewed with Acrobat Reader for the animated videos.}}}
	\vspace{-0.15in}
	\label{fig:fine_grained}
\end{figure*}

\section{Experiments}
\subsection{Experimental Settings}

\begin{table}[t]
	\centering
	\caption{Fine-grained motion control results on WebVid-10M.}
	\label{tab:webvid}
	\vspace{-5pt}
	\setlength{\tabcolsep}{0.4cm}
	\resizebox{\linewidth}{!}{
		\begin{tabular}{lccc}
				\toprule
				\textbf{Approach} & \textbf{FVD} ($\downarrow$) & \textbf{FID} ($\downarrow$) & \textbf{Frame Consis.} ($\uparrow$) \\
				\midrule
				DragNUWA \cite{2023dragnvwa} & 96.65 & 13.19 & 0.9888 \\
				MOFA-Video \cite{2024mofa} & \underline{87.70} & \underline{12.18} & \underline{0.9894} \\ 
				MotionPro & \textbf{59.88} & \textbf{10.40} & \textbf{0.9895} \\ 
				\bottomrule
	\end{tabular}}
	\vspace{-15pt}
\end{table}

\textbf{Benchmarks.} 
We empirically verify the merit of MotionPro on two benchmarks, i.e., WebVid-10M~\cite{2021Webvid} and our proposed MC-Bench.
The \textbf{WebVid-10M} dataset consists of $10.7M$ video-caption pairs.
There are $5K$ videos in the validation set and we sample $1K$ videos for evaluation.
For each video, trajectories sampled at a ratio of $15\%$ along with the first frame serve as the input condition for fine-grained I2V motion generation.
We follow the protocols in recent controllable I2V advance~\cite{2024mofa} and choose the Frechet Video Distance (FVD)~\cite{2018fvd}, Frechet Image Distance (FID)~\cite{2017gans}, and Frame Consistency (Frame Consis.)~\cite{2023fatezero} of CLIP~\cite{2021clip} features as the evaluation metrics on WebVid-10M.

In practical applications, users typically prefer to control video generation through a limited number of representative trajectories, often just one or two.
The automatically sampled trajectories employed in WebVid-10M do not adequately represent this scenario, thereby potentially compromising the validity of the evaluation.
Thus, we introduce \textbf{MC-Bench}, a new benchmark with $1.1K$ reference images and user-annotated trajectories, which is tailored for the evaluation of controllable I2V generation.
More details about the new dataset are provided in the supplementary material.
Due to the absence of ground-truth video, FVD and FID metrics are not applicable to MC-Bench.
In addition to Frame Consistency, we utilize the Mean Distance (MD) to measure the alignment between generated motion and input trajectory.
Two evaluation protocols are exploited for this target, i.e., MD-Img and MD-Vid.
MD-Img is proposed by DragDiffusion~\cite{2024dragdiffusion} which estimates the frame-level mean Euclidean distance between trajectories of input and generated frames.
To further validate the video-level trajectory accuracy via MD-Vid, we replace the image correspondence detection model DIFT~\cite{2023dift} in MD-Img with the video tracking model CoTracker~\cite{2023cotracker}, which supplies a more precise trajectory reference.

\begin{table}[t]
	\centering
	\caption{Fine-grained motion control results on MC-Bench.}
	\label{tab:fine-grained}
	\vspace{-5pt}
	\setlength{\tabcolsep}{0.25cm}
	\resizebox{\linewidth}{!}{
		\begin{tabular}{lccc}
			\toprule
			\textbf{Approach} & \textbf{MD-Img} ($\downarrow$)~ & ~\textbf{MD-Vid} ($\downarrow$)~~ & \textbf{Frame Consis.} ($\uparrow$) \\
			\midrule
			DragDiffusion \cite{2024dragdiffusion} & 14.70 & 13.84 & 0.9947  \\
			MOFA-Video \cite{2024mofa} & \underline{13.94} & \underline{10.50} & \textbf{0.9972}  \\ 
			MotionPro & \textbf{10.56} & \textbf{8.34} & $\underline{0.9962}$ \\ 
			\bottomrule
	\end{tabular}}
	\vspace{-0.2in}
\end{table}

\textbf{Implementation Details.} 
In MotionPro, we employ SVD~\cite{svd2023} as our base architecture.
Each training sample is $16$-frames video clip and the sampling rate is $8$ fps.
We fix the resolution of each frame as $320\times 512$, which is centrally cropped from the resized video.
The local region size $k$ is set as $8$ and the minimal mask ratio $r_{min}$ is set as $0.95$ determined by cross validation.
We set the rank of LoRA parameters as $32$.
The motion encoder and LoRA layers are trained via AdamW optimizer with the base learning rate $1\times 10^{-5}$. 
All experiments are conducted on 6 NVIDIA A800 GPUs with minibatch size $48$.

\subsection{Evaluation on Fine-grained Motion Control}
We first evaluate MotionPro on the fine-grained motion control for I2V generation.
The performances on WebVid-10M and MC-Bench are summarized in Table~\ref{tab:webvid} and Table~\ref{tab:fine-grained}, respectively.
Our MotionPro consistently achieves better performances on WebVid-10M across different metrics.   
In particular, MotionPro attains the FVD of $59.88$, outperforming the best competitor MOFA-Video by $27.82$.
The better FVD indicates the better alignment of data distribution between the generated and ground-truth videos.
Such results basically verify the superiority of exploring precise region-wise trajectory to strengthen fine-grained motion dynamic learning.
On MC-Bench, MotionPro leads to performance boosts against baselines in terms of MD-Img and MD-Vid, showing better alignment between the user input trajectory and synthesized videos.
Note that MOFA-Video exploits a two-stage controllable I2V framework that first densifies the input trajectories through conditional motion propagation (CMP), and then calibrates video diffusion process using the estimated dense trajectories.
In contrast, MotionPro learns precise motion patterns by directly referring region-wise trajectory via adaptive feature modulation, thus enhancing the motion-trajectory alignment, as evidenced by the better MD-Img and MD-Vid performances.
Besides, the CMP technique in MOFA-Video generally focuses on flow completion in the local region surrounding the input trajectory while neglecting potential movements in other areas.
\begin{table}[t]
	\centering
	\caption{Object-level motion control results on MC-Bench.}
	\label{tab:object}
	\vspace{-5pt}
	\setlength{\tabcolsep}{0.3cm}
	\resizebox{\linewidth}{!}{
		\begin{tabular}{lccc}
			\toprule
			\textbf{Approach} & \textbf{MD-Img} ($\downarrow$) & \textbf{MD-Vid} ($\downarrow$) & \textbf{Frame Consis.} ($\uparrow$) \\
			\midrule
			MOFA-Video \cite{2024mofa} & 15.56 & 12.04 & \textbf{0.9951}  \\
			DragAnything \cite{2024draganything} & \underline{12.30} & \underline{11.37} & 0.9917  \\ 
			MotionPro & \textbf{10.48} & \textbf{8.59} & $\underline{0.9943}$ \\ 
			\bottomrule
	\end{tabular}}
	\vspace{-0.3in}
\end{table}
Thus, MOFA-Video tends to synthesize videos with less motion dynamics and obtains slightly higher Frame Consistency (approximately $0.001$).
To substantiate this, we calculate the average flow magnitude of videos generated by MOFA-Video, which achieves $4.95$.
In comparison, MotionPro attains a higher value of $8.95$, verifying that our model achieves greater motion variability while maintaining better motion-trajectory alignment.

\begin{figure*}[t]
	\centering
	\begin{tabularx}{0.85\textwidth}{XXXX}
		\centering \textbf{Input Control} & \centering \textbf{MOFA-Video} & \centering \textbf{DragAnything} & \centering \textbf{MotionPro}
	\end{tabularx}
	\resizebox{0.85\textwidth}{!}{
		\begin{tabular}{
				@{}
				p{\dimexpr(\textwidth-0.3em*3)/4} @{\hspace{0.3em}}
				p{\dimexpr(\textwidth-0.3em*3)/4} @{\hspace{0.3em}}
				p{\dimexpr(\textwidth-0.3em*3)/4} @{\hspace{0.3em}}
				p{\dimexpr(\textwidth-0.3em*3)/4} @{}}
			\includegraphics[width=\linewidth]{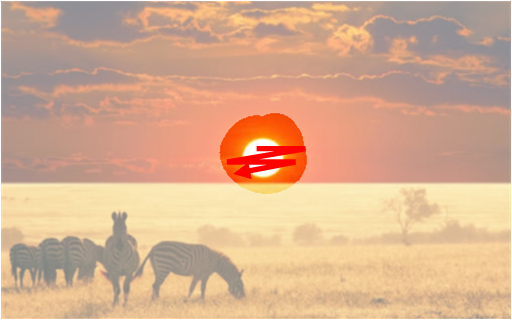} & 
			\animategraphics[autoplay,loop,width=\linewidth]{8}{videos/obj_case_final/mofa/case1/pngs/}{01}{16} & 
			\animategraphics[autoplay,loop,width=\linewidth]{8}{videos/obj_case_final/draganything/case1/pngs/}{01}{16} &
			\animategraphics[autoplay,loop,width=\linewidth]{8}{videos/obj_case_final/ours/case1/pngs/}{01}{16} \\
			
			\includegraphics[width=\linewidth]{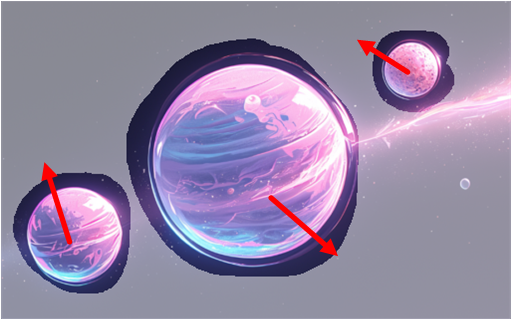} & 
			\animategraphics[autoplay,loop,width=\linewidth]{8}{videos/obj_case_final/mofa/case2/pngs/}{01}{16}     & 
			\animategraphics[autoplay,loop,width=\linewidth]{8}{videos/obj_case_final/draganything/case2/pngs/}{01}{16} &
			\animategraphics[autoplay,loop,width=\linewidth]{8}{videos/obj_case_final/ours/case2/pngs/}{01}{16} 
			\\
			
			\includegraphics[width=\linewidth]{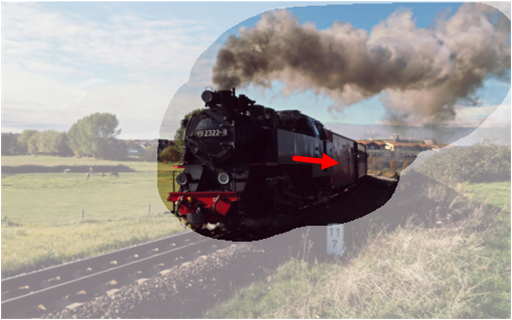} & 
			\animategraphics[autoplay,loop,width=\linewidth]{8}{videos/obj_case_final/mofa/case3/pngs/}{01}{16}     & 
			\animategraphics[autoplay,loop,width=\linewidth]{8}{videos/obj_case_final/draganything/case3/pngs/}{01}{16} &
			\animategraphics[autoplay,loop,width=\linewidth]{8}{videos/obj_case_final/ours/case3/pngs/}{01}{16}
			\\
			
			\includegraphics[width=\linewidth]{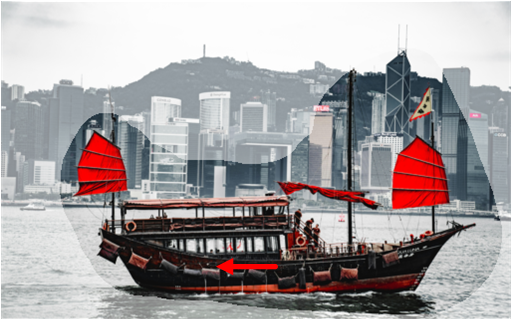} & 
			\animategraphics[autoplay,loop,width=\linewidth]{8}{videos/obj_case_final/mofa/case4/pngs/}{01}{16}     & 
			\animategraphics[autoplay,loop,width=\linewidth]{8}{videos/obj_case_final/draganything/case4/pngs/}{01}{16} &
			\animategraphics[autoplay,loop,width=\linewidth]{8}{videos/obj_case_final/ours/case4/pngs/}{01}{16}  
	\end{tabular}}
	\vspace{-0.08in}
	\caption{Examples of object-level motion control results on MC-Bench. The input control signals include reference image, trajectory and motion mask. MotionPro can successfully handle complicated (e.g., the round trip of sun in the 1st case) and counterintuitive (e.g., the train moving back in the 3rd case) motion-trajectory alignment. \textit{\textbf{Best viewed with Acrobat Reader.}}}
	\label{fig:object_case}
	\vspace{-0.06in}
\end{figure*}

\begin{figure*}[t]
	\centering
	\includegraphics[width=0.99\linewidth]{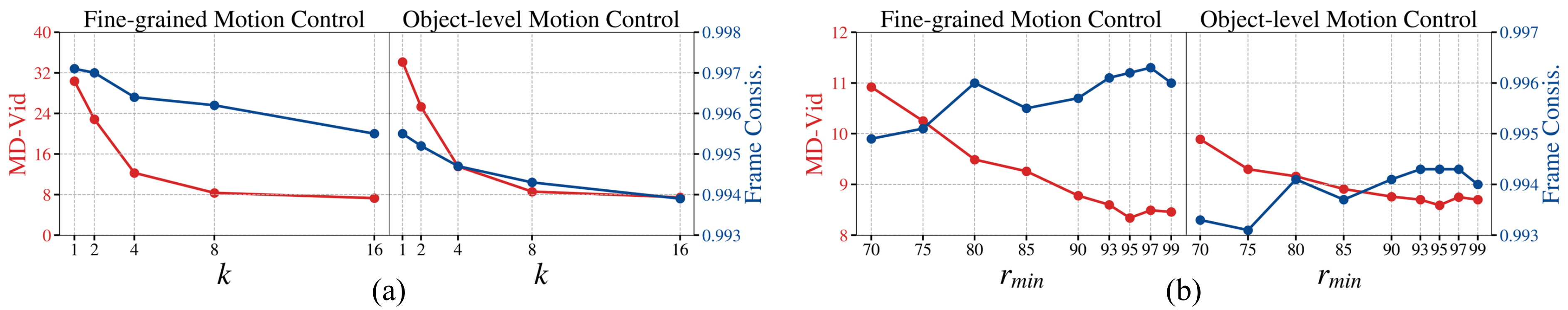}
	\vspace{-0.1in}
	\caption{Performance comparisons of MD-Vid and Frame Consistency on MC-Bench under the settings of both fine-grained and object-level motion control by using different (a) local region size $k$ and (b) minimal mask ratio $r_{min}$ in MotionPro.}
	\label{fig:ablation}
	\vspace{-0.18in}	
\end{figure*}

\begin{figure*}[t]
	\centering
	\begin{tabularx}{\textwidth}{XXXXXXXX}
		\centering Control & \centering k=16 & \centering \textbf{k=8} & \centering Control & \centering k=1 & \centering k=2 & \centering k=4 & \centering \textbf{k=8}
	\end{tabularx}
	\vspace{-0.8em}
	\begin{tabular}{
			@{}
			p{\dimexpr(\textwidth-0.1em*7)/8} @{\hspace{0.1em}}
			p{\dimexpr(\textwidth-0.1em*7)/8} @{\hspace{0.1em}}
			p{\dimexpr(\textwidth-0.1em*7)/8} @{\hspace{0.1em}}
			p{\dimexpr(\textwidth-0.1em*7)/8} @{\hspace{0.1em}}
			p{\dimexpr(\textwidth-0.1em*7)/8} @{\hspace{0.1em}}
			p{\dimexpr(\textwidth-0.1em*7)/8} @{\hspace{0.1em}}
			p{\dimexpr(\textwidth-0.1em*7)/8} @{\hspace{0.1em}}
			p{\dimexpr(\textwidth-0.1em*7)/8} @{}}
		\includegraphics[width=\linewidth]{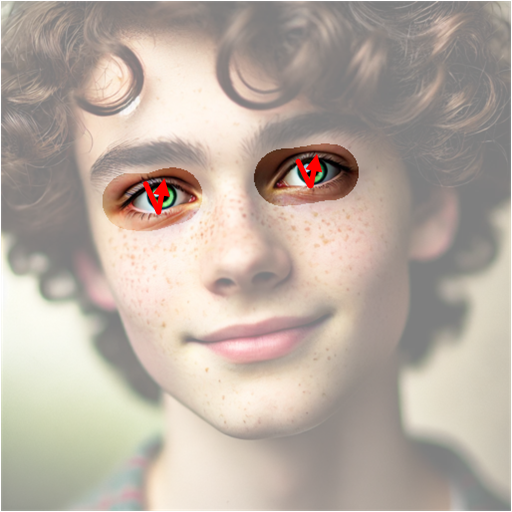} & 
		\animategraphics[autoplay,loop,width=\linewidth]{8}{videos/block_size_final/size_16/case1/pngs/}{01}{16} & 
		\animategraphics[autoplay,loop,width=\linewidth]{8}{videos/block_size_final/size_8/case1/pngs/}{01}{16} &
		\includegraphics[width=\linewidth]{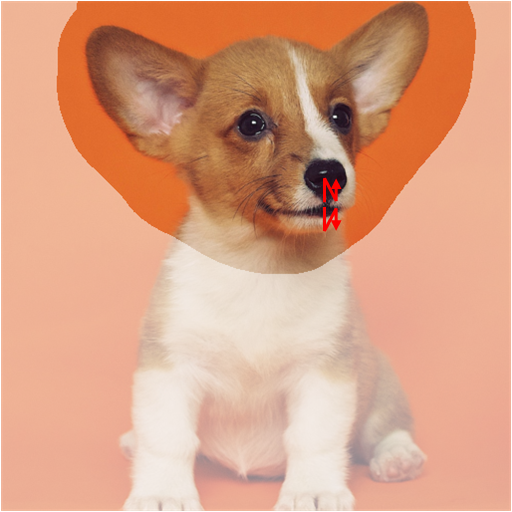} & 
		\animategraphics[autoplay,loop,width=\linewidth]{8}{videos/block_size_final/size_1/case2/pngs/}{01}{16} & 
		\animategraphics[autoplay,loop,width=\linewidth]{8}{videos/block_size_final/size_2/case2/pngs/}{01}{16}     & 
		\animategraphics[autoplay,loop,width=\linewidth]{8}{videos/block_size_final/size_4/case2/pngs/}{01}{16} & 
		\animategraphics[autoplay,loop,width=\linewidth]{8}{videos/block_size_final/size_8/case2/pngs/}{01}{16}
	\end{tabular}
	\vspace{-0.06in}
	\caption{Visualization of controllable I2V generation results with different local region size $k$ in MotionPro. \textit{\textbf{Viewed with Acrobat Reader.}}}
	\label{fig:ablation_block_size}
	\vspace{-0.22in}	
\end{figure*}

Figure~\ref{fig:fine_grained} further showcases three I2V generation results controlled by the user input trajectory and region mask on MC-Bench.
Generally, the videos synthesized by our MotionPro exhibits more natural movement and better alignment with input trajectory than the baseline methods.
For instance, DragNUWA suffers from motion misinterpretation issue which wrongly generates videos with camera movement instead of object moving (e.g., the 1st and 2nd cases).
The videos generated by MOFA-Video usually present unnatural object movement with local part distortion, e.g., the nose of raccoon in the 2nd case.
We speculate that such distortion is caused by the lack of global region guidance in MOFA-Video, where the region mask is only employed for flow masking as post-processing.
Our MotionPro, in comparison, integrates the information of motion mask into 3D-UNet on the fly to facilitate the modeling of holistic motion correlation.
Thus, the synthesized videos by MotionPro reflect more rational fine-grained movement.

\subsection{Evaluation on Object-level Motion Control} 
Next, we conduct evaluation on object-level motion control for I2V generation.
Table~\ref{tab:object} lists the performances of different approaches on MC-Bench.
Overall, MotionPro attains the best performances on the metrics of MD-Img and MD-Vid.
Specifically, MotionPro obtains $10.48$ of MD-Img and $8.59$ of MD-Vid, reducing the Mean Distance of the best competitor DragAnything by $1.82$ and $2.78$, respectively.
The improvements again confirm the merit of leveraging the duet of region-wise trajectory and motion mask for precise motion control.
Similar performance trend on Frame Consistency can be also observed in the table.

Figure~\ref{fig:object_case} shows the visual comparison of four object-level motion control results by using different approaches on MC-Bench.
Compared to other baselines, videos generated by MotionPro can precisely match the input trajectory and maintain natural object-level motion dynamics.
MOFA-Video still faces the challenge of local part distortion (e.g., only the train rear moving back in the 3rd case) and video generation with limited motion dynamics (e.g., the 4th case).
Though DragAnything effectively aligns pixel movement with the input trajectory, certain instances (e.g., the 1st and 4th cases) misinterpret the trajectory as camera motion rather than object movement.
In contrast, MotionPro nicely capitalizes on trajectory information to refine video denoising, and specifies motion category with the region mask, endowing images with better object motion.

\subsection{Ablation Study on MotionPro}
In this section, we perform ablation study to delve into the design of MotionPro for controllable I2V generation. Here, all experiments are conducted on MC-Bench.

\textbf{Local Region Size.} 
We first investigate the choice of local region size $k$ for region-wise trajectory design in our MotionPro.
Figure~\ref{fig:ablation}(a) compares the performances of MD-Vid and Frame Consistency on both fine-grained and object-level motion control by using different $k$.
The variation of Frame Consistency is minor (less than $0.01$) across different settings, and the MD-Vid decreases when using larger $k$.
When $k$ is small (e.g., $1$ or $2$), the kept trajectories are less in each local region and the control signals are weaken for motion control, leading to the inferior trajectory matching performance.
Meanwhile, the improvement of MD-Vid is marginal when increasing $k$ to $16$.
Specifically, using large $k$ will extend the input trajectory over a large region, which affects the fine-grained motion control.
Accordingly, we exploit $k=8$ to extract the region-wise trajectory as the motion condition.
Figure~\ref{fig:ablation_block_size} further illustrates the I2V generation results with different $k$.
As shown in this figure, the synthesized videos with $k=8$ present more natural motion dynamics and more precise motion-trajectory alignment.
Moreover, the unnatural fine-grained motion as shown in the case when $k=16$ validates our analysis on the influence of overlarge region size.

\textbf{Minimal Mask Ratio.}
To explore the effect of minimal mask ratio $r_{min}$ in trajectory selection stage, we then measure the motion control performance by conducting different $r_{min}$ in Figure~\ref{fig:ablation}(b).
Overall, Frame Consistency is not sensitive when changing $r_{min}$ on both fine-grained and object-level motion control settings.
Meanwhile, the performance of MD-Vid becomes better with the increase of the mask ratio at the beginning.
The results are expected since using small value of $r_{min}$ will sample more trajectories for model training, which enlarges the gap between training and real-world inference (i.e., only using one or two trajectories).
Conversely, employing a large value of mask ratio (e.g., $0.99$) could make it difficult to optimize networks with scarce trajectory signals.
Thus, we empirically set $r_{min}$ as $0.95$ to obtain the best motion-trajectory alignment in the generated videos.

\begin{figure}
	\centering
	\vspace{-0.1in}
	\includegraphics[width=0.6\linewidth]{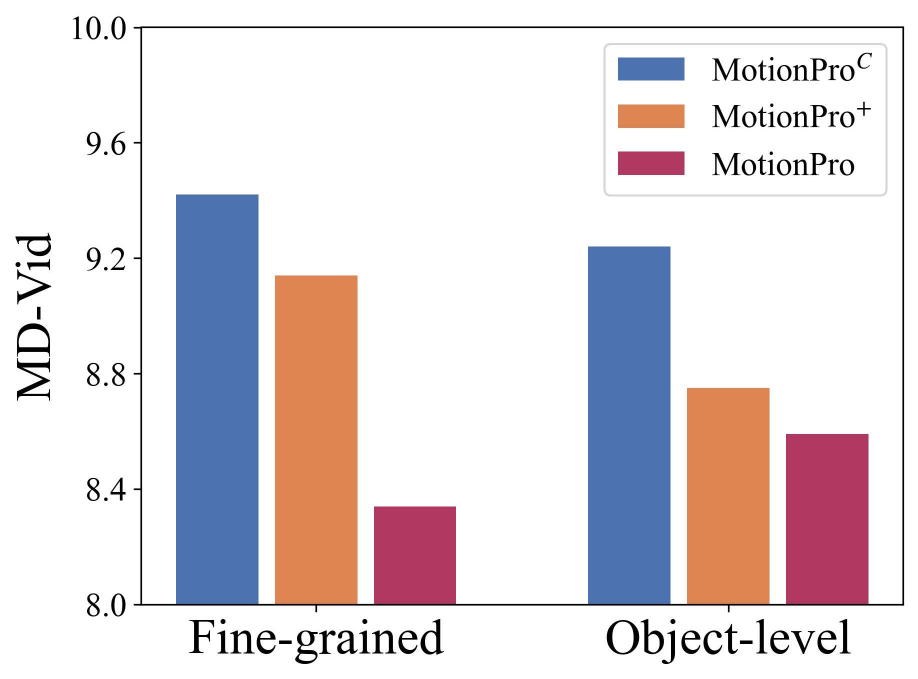}
	\vspace{-0.13in}
	\caption{MD-Vid ($\downarrow$) among different multi-scale feature injection approaches on MC-Bench.}
	\vspace{-0.20in}
	\label{fig:ab_modulation}
\end{figure}

\textbf{Multi-scale Feature Injection.}
We also investigate different multi-scale feature injection strategies in MotionPro. 
Figure~\ref{fig:ab_modulation} details the MD-Vid performance comparisons among different variants of our MotionPro.
\textbf{MotionPro}$^{C}$ concatenates the multi-scale features learnt by motion encoder with the video latent features along channel dimension in each scale.
\textbf{MotionPro}$^{+}$ replaces the channel-wise feature concatenation in {MotionPro}$^{C}$ with the feature summation.
In comparison, our proposal (\textbf{MotionPro}) injects the control signals into 3D-UNet via the adaptive feature modulation.
Overall, MotionPro exhibits better MD-Vid performances against other two variants.
In direct feature aggregation methods such as concatenation or summation, information exchange requires strict spatial-temporal alignment between each other. 
In contrast, there is no such requirement for feature modulation, as it indirectly utilizes estimated scale and bias for feature regulation. 
Consequently, such feature injection demonstrates enhanced capacity to extract relevant information from input signals, potentially leading to improved motion control performance.

\section{Conclusions}
This paper explores the motion condition formulation and the motion-trajectory alignment in diffusion models for controllable I2V generation.
In particular, we study the problem from the viewpoint of integrating accurate motion control signals into video denoising to regulate motion generation.
To materialize our idea, we have devised MotionPro, which leverages the region-wise trajectory and motion mask as the condition to calibrate video generation in a feature modulation manner.
%The region-wise trajectory preserves the original optical flow information in each local region, characterizing the fine-grained motion details.
The region-wise trajectory directly exploits the original trajectory information in each local region, characterizing more accurate motion details.
The motion mask derived from the optical flow maps presents holistic motion and aims to identify exact motion category.
The collaboration of the two signals regulates video denoising for natural motion synthesis with precise motion-trajectory alignment. 
Moreover, we have carefully construct a new benchmark, i.e., MC-Bench, with $1.1K$ user-annotated image-trajectory pairs for both fine-grained and object-level motion control evaluation.
Extensive experiments on WebVid-10M and MC-Bench validate the superiority of our proposal over state-of-the-art approaches.

\textbf{Acknowledgments.} This work was supported in part by the Beijing Municipal Science and Technology Project No.Z241100001324002 and Beijing Nova Program No.20240484681.

{
\small
\bibliographystyle{ieeenat_fullname}
\bibliography{main}
}


\clearpage
\section{Supplementary Material}

\maketitle

The supplementary material contains: 1) the dataset details of MC-Bench; 2) baseline choices and experimental details; 3) the human evaluation of motion control; 4) robustness of motion mask; 5) the application of camera control; 6) runtime comparison; 7) ablation on control signals.

\subsection{Dataset Details of MC-Bench}
\begin{figure*}[t]
	\centering
	\includegraphics[width=0.99\linewidth]{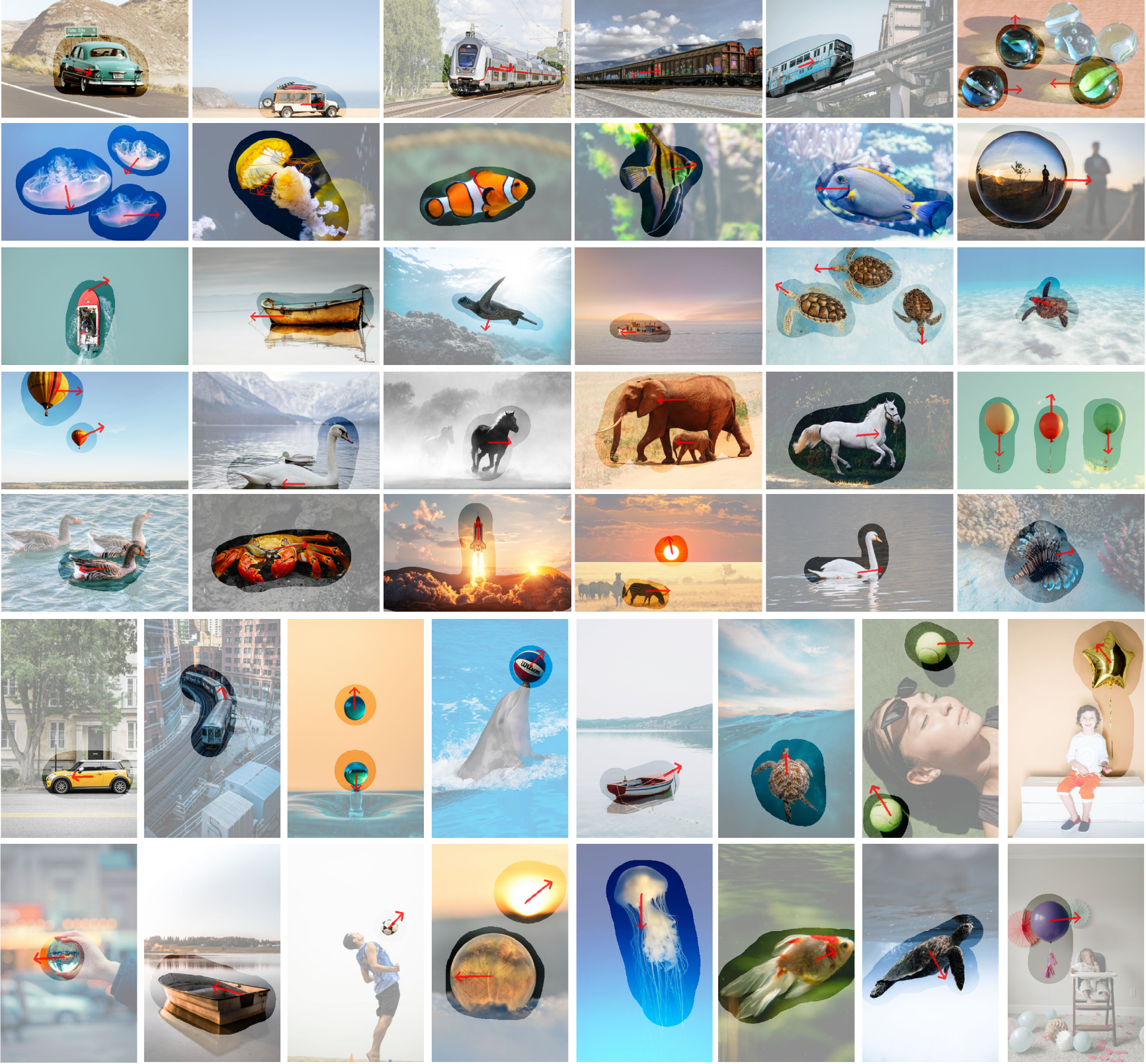}
	\vspace{-0.05in}
	\caption{Visual examples from MC-Bench for object-level motion control evaluation. Each reference image is annotated with trajectory and motion mask for image-to-video generation.}
	\label{fig:obj_data}
	\vspace{-0.1in}	
\end{figure*}

\begin{figure*}[t]
	\vspace*{-2in}  
	\centering
	\includegraphics[width=0.99\linewidth]{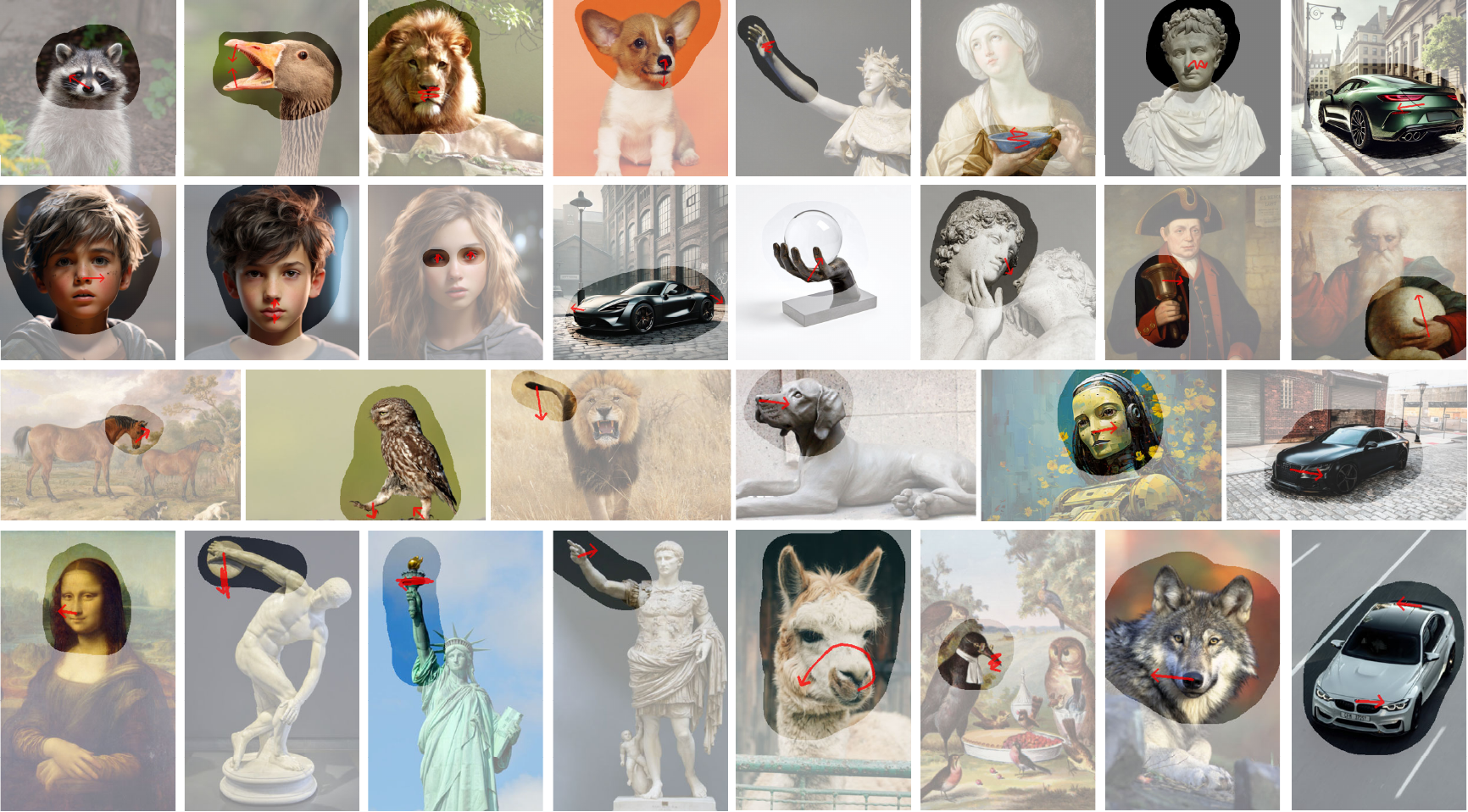}
	\vspace{-0.05in}
	\caption{Visual examples from MC-Bench for fine-grained motion control evaluation. Each reference image is annotated with trajectory and motion mask for image-to-video generation.}
	\label{fig:fine_data}
	\vspace{-0.1in}	
\end{figure*}

The proposed MC-Bench consists of $412$ high-quality reference images and corresponding $1.1K$ user-annotated trajectories. 
We collect the reference images with different visual contents, including animal, human, vehicle, etc.
There are $72$ images sampled from the public DragBench~\cite{2024dragdiffusion} and we further extend it with $340$ additional images.
Specifically, all the self-collected images about human are automatically generated by DALL·E3~\cite{dalle3} to avoid the potential legal concerns.
The remaining self-collected images are real photos which are first crawled on the Pexels platform and then filtered according to the visual quality.
For each reference image, the annotator is required to brush the motion region and draw the movement trajectory according to user intention (i.e., fine-grained local part moving or global object moving).
During trajectory annotation, all annotators are encouraged to ensure the trajectory diversity, including some complicated trajectories.
Finally, the benchmark is annotated with $460$ image-trajectory pairs for fine-grained motion control evaluation, and $680$ image-trajectory pairs for object-level motion control evaluation, respectively.
Figure~\ref{fig:fine_data} and Figure~\ref{fig:obj_data} further illustrate several visual examples (reference image, trajectory and motion mask) from MC-Bench for the two evaluations.

\subsection{Baseline Choices and Experimental Details}
For the evaluation on WebVid-10M~\cite{2021Webvid} of fine-grained motion control, we adopt the commonly-used protocol in recent controllable image-to-video (I2V) advance~\cite{2024mofa}.
Specifically, for each video, we sample the optical flow at the ratio of $15\%$ as the sparse trajectories, which are combined with the first frame as the input condition.
Under this experimental setting, we choose DragNUWA~\cite{2023dragnvwa} and MOFA-Video~\cite{2024mofa} as baselines for comparison.
Notably, DragAnything~\cite{2024draganything} is deliberately designed for object-level motion control, which only accepts a single trajectory of object, making it inapplicable for fine-grained motion control. 
Therefore, DragAnything is not involved for comparison in this setting.

For the fine-grained motion control on MC-Bench, we compare our MotionPro with DragDiffusion~\cite{2024dragdiffusion} and MOFA-Video.
DragNUWA is not included in this comparison since it only relies on trajectories and lacks the input of motion regions. 
Thus, DragNUWA usually suffers from the misinterpretation of object and camera movement, making the comparison unfair.
The baseline of DragDiffusion is a recent trajectory-guided image editing advance, which also offers convincing results for comparison.
To adapt DragDiffusion for video generation, we divide the input trajectories into 15 segments and independently feed each segment into DragDiffusion to generate target frame.
All the synthesized frames are concatenated as the final video.

In the evaluation of object-level motion control on MC-Bench, both MOFA-Video and DragAnything are employed as baselines for performance comparison.
To facilitate DragAnything in disentangling object and camera moving, we add static points in regions outside the motion mask areas to help DragAnything generate object-level motion instead of camera moving for evaluation.
It's worth to noting that MotionPro learns object and camera motion control on ``in-the-wild'' video data (e.g., WebVid-10M) without applying special data filtering.

\begin{table*}
	\centering
	\caption{Human evaluation of user preference ratios (\%) over both fine-grained and object-level motion control on MC-Bench.}
	\vspace{-0.05in}
	\setlength{\tabcolsep}{0.3cm}\resizebox{1\textwidth}{!}{
		\begin{tabular}{lccc| ccc}
			\toprule
			\multicolumn{1}{l}{\multirow{2}{*}{\textbf{Evaluation Items}}} 
			& \multicolumn{3}{c}{\textbf{Fine-grained Motion Control}} 
			& \multicolumn{3}{c}{\textbf{Object-level Motion Control}} \\ 
			\cmidrule(lr){2-4} \cmidrule(lr){5-7} 
			& DragDiffusion~\cite{2024dragdiffusion} & MOFA-Video~\cite{2024mofa} & MotionPro
			& MOFA-Video~\cite{2024mofa} & DragAnything~\cite{2024draganything} & MotionPro \\ 
			\midrule
			Motion Quality ($\uparrow$) & 3.12 & $\underline{21.88}$ & \textbf{75.00} & 12.50 & $\underline{18.75}$ & \textbf{68.75} \\ 
			Temporal Coherence ($\uparrow$) & 6.25 & $\underline{40.63}$ & \textbf{53.12} & $\underline{25.00}$ & 15.63 & \textbf{59.37} \\  
			Trajectory Alignment ($\uparrow$) & 9.37 & $\underline{18.75}$ & \textbf{71.88} & 15.62 & $\underline{21.88}$ & \textbf{62.50} \\ 
			\bottomrule
		\end{tabular}
	}
	\label{tab:human}
\end{table*}

\begin{figure*}
	\centering
	\includegraphics[width=1.0\linewidth]{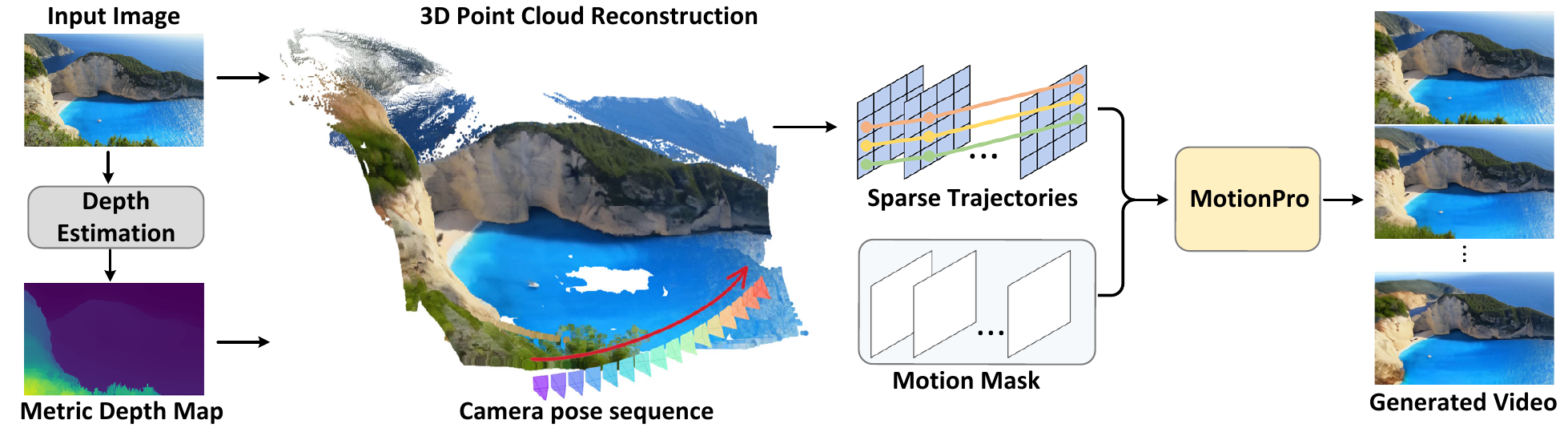}
	\caption{An illustration of I2V camera control using the condition of camera pose sequence in our MotionPro.}
	\label{fig:camera}
\end{figure*}

\subsection{Human Evaluation}
In addition to the evaluation over automatic metrics, we also conduct human evaluation to investigate user preferences from three perspectives (i.e., motion quality, temporal coherence and trajectory alignment) across different controllable I2V approaches.
In particular, we randomly sample 200 generated videos from both fine-grained and object-level motion control for evaluation. 
Through the Amazon MTurk platform, we invite 32 evaluators, and ask each evaluator to choose the best one from the generated videos by all models given the same inputs.

Table~\ref{tab:human} shows the user preference ratios across different models on MC-Bench.
Overall, our MotionPro clearly outperforms all baselines in terms of the three criteria on both fine-grained and object-level motion control.
The results demonstrate the advantage of leveraging complementary region-wise trajectory and motion mask to benefit video synthesis with natural motion, desirable temporal coherence and precise motion-trajectory alignment.

\subsection{Ablation on control signals.}We also include two runs (MotionPro$^{-}_{traj}$: replaces region-wise trajectory with random trajectory, MotionPro$^{-}_{mask}$: disables motion mask with all-one masks).
Their FVD ($73.7$ and $66.2$) on WebVid-10M are inferior to our MotionPro ($59.88$), which validates the effectiveness of our two control signal designs for precise motion formulation.

\subsection{Robustness of motion mask}
To be clear, motion mask in our MotionPro refers to the rough dynamic region and does not require precisely-aligned shape at inference.
We show I2V results controlled by the same trajectory with various motion masks in Figure \ref{fig:mask}, which show strong robustness.
Such generalization merit is attributed to the use of estimated motion mask (flow map estimated by DOT) at training, rather than ground-truth precise motion mask.

\subsection{Application: Camera Control}
Our learnt MotionPro naturally supports two applications of camera control without additional training.
The first application is controlling object and camera motion simultaneously with multiple trajectories in I2V generation.
Another application is the I2V camera control by exploiting the sequence of camera poses as input condition. To be clear, motion mask in our MotionPro refers to the rough dynamic region and does not require precisely-aligned shape at inference.

\textbf{Simultaneous object and camera motion control.}
In this setting, we simply set the input motion mask as all-ones matrix, and feed multiple trajectories that reflect the object and background moving into MotionPro for I2V generation.
The video cases are provided in the offline project page.

\textbf{Camera control with camera poses.}
Figure~\ref{fig:camera} illustrates the process of camera control using the condition of camera pose sequence in MotionPro.
Concurrently, given an input image and the camera pose sequence, we first estimate the metric depth map of the image using ZoeDepth~\cite{2023ZoeDepth}.
Next, we lift the 2D pixels to 3D point cloud using the metric depth map.
Through projecting the point cloud into 2D space given the camera pose, we can determine the corresponding 2D positions of the same 3D points under the new view.
By calculating the 2D displacement of the pixels projected from the same 3D points in the original and new views, the camera pose sequence is then converted into the sparse trajectories.
Finally, we feed the sparse trajectories and all-ones motion mask into MotionPro for I2V synthesis.
The video cases are provided in the offline project page.

\subsection{Runtime Comparison}
For 16-frame video generation (resolution: $512\times320$, on single NVIDIA H100 GPU), the runtime of MotionPro is 17 sec, which is comparable to baselines (DragNUWA: 27, DragDiffusion: 320, MOFA-Video: 15, DragAnything: 32).

\begin{figure*}[t]
	\centering
	\animategraphics[autoplay,loop,width=0.98\linewidth]{8}{videos/mask_area/}{0001}{0016}
	\vspace{-0.1in}
	\caption{MotionPro conditioned on diverse mask shapes. \textit{\textbf{Best viewed with Acrobat Reader.}}}
	\label{fig:mask}
\end{figure*}

\end{document}